\documentclass{preprint}
\usepackage[numbers,comma,sort&compress]{natbib}
\usepackage{graphicx}%
\usepackage{amsmath,amssymb,amsfonts}%
\usepackage{xcolor}%
\usepackage{booktabs}%
\usepackage{makecell}
\usepackage{comment}
\usepackage{tabularx}
\usepackage{longtable}
\usepackage{pdflscape}
\usepackage[colorinlistoftodos]{todonotes}
\usepackage{multirow}
\usepackage[colorlinks=true]{hyperref}
\usepackage{tikz}
\usepackage{forest}
\usetikzlibrary{trees,positioning,shapes,shadows,arrows.meta}
\usepackage{nopageno}
\usepackage{colortbl}
\usepackage{tabularray}
\usepackage{xltabular}

\begin{document}

\title{Deep learning with noisy labels in medical prediction problems: a scoping review}

\author[1,3]{Yishu Wei, Ph.D.}
\author[2]{Yu Deng, Ph.D.}
\author[1]{Cong Sun, Ph.D.}
\author[1]{Mingquan Lin, Ph.D.}
\author[4]{Hongmei Jiang, Ph.D.}
\author[1,*]{Yifan Peng, Ph.D.}

\affil[1]{Department of Population Health Sciences, Weill Cornell Medicine, New York, 10065, US}
\affil[2]{Statistical Innovation Group, Data \& Statistical Science, AbbVie Inc., Chicago, 60064, US}
\affil[3]{Reddit Inc., San Francisco, 16093, US}
\affil[4]{Department of Statistics and Data Science, Northwestern University, Evanston, 60208, US}

\affil[*]{Corresponding author(s): yip4002@med.cornell.edu}

\maketitle

\begin{abstract}
\textbf{Objectives}: Medical research faces substantial challenges from noisy labels attributed to factors like inter-expert variability and machine-extracted labels. Despite this, the adoption of label noise management remains limited, and label noise is largely ignored. To this end, there is a critical need to conduct a scoping review focusing on the problem space. This scoping review aims to comprehensively review label noise management in deep learning-based medical prediction problems, which includes label noise detection, label noise handling, and evaluation. Research involving label uncertainty is also included.

\textbf{Methods}: Our scoping review follows the Preferred Reporting Items for Systematic Reviews and Meta-Analyses (PRISMA) guidelines. We searched 4 databases, including PubMed, IEEE Xplore, Google Scholar, and Semantic Scholar. Our search terms include ``noisy label AND medical / healthcare / clinical'', ``uncertainty AND medical / healthcare / clinical'', and ``noise AND medical / healthcare / clinical''.

\textbf{Results}: A total of 60 papers met inclusion criteria between 2016 and 2023. A series of practical questions in medical research are investigated. These include the sources of label noise, the impact of label noise, the detection of label noise, label noise handling techniques, and their evaluation. Categorization of both label noise detection methods and handling techniques are provided.

\textbf{Discussion}: From a methodological perspective, we observe that the medical community has been up to date with the broader deep-learning community, given that most techniques have been evaluated on medical data. We recommend considering label noise as a standard element in medical research, even if it is not dedicated to handling noisy labels. Initial experiments can start with easy-to-implement methods, such as noise-robust loss functions, weighting, and curriculum learning.


\end{abstract}

\section{Introduction}

Numerous studies have shown the extensive use of deep learning (DL) across various medical domains. Remarkably, DL has achieved performance on par with domain experts in specific applications. However, the performance of DL training dramatically relies upon the quality of the underlying data, which can be heavily impacted by data noise. This noise can be categorized into two categories: feature and label noise. In this review, we focus on the latter.

The medical research landscape presents unique characteristics related to label noise. Firstly, many disease diagnoses are challenging even for experts, leading to diverse outcomes~\cite{wallace2008agreement}. Moreover, there is a prevalent issue with the scarcity of expert-annotated labels, owing to the costly and time-consuming nature of the annotation process. To mitigate this, many researchers use labels generated through natural language processing (NLP), crowd-sourcing, or semi-supervised pseudo-label approaches. These alternatives, however, often lead to considerable label noises. Despite the significant presence of noisy labels in medical data, this issue hasn't received much attention in studies focused on medical machine learning.

To bridge this gap, this scoping review is intended to address several pertinent questions: the sources of label noise, the consequential effects of label noise, methods for identifying and handling label noise, as well as their evaluation. Our target readers are researchers and practitioners engaged in applying deep learning within the medical domain. Through examining these questions, we aim to highlight the challenge posed by noisy labels and encourage the adoption of suitable measures to mitigate this issue.

There are several reviews regarding learning from noisy labels in the general machine-learning domain. For example, Frenay et al. covered studies before the advent of deep learning~\cite{frenay2014comprehensive}. 
Song et al. focused on model architecture change to handle label noise~\cite{Song2023learning}. 
Algan et al. comprehensively reviewed noisy label handling methodology~\cite{algan2021image}.
Liang et al. summarized methods regarding sample selection and loss correction~\cite{liang2022review}.
Among all the reviews currently available, we only came across one survey in the medical domain~\cite{karimi2020deep}.

Our review offers several extensions to previous studies. First, previous studies focused on methodologies before 2018 within the broader deep-learning community and cited limited references in the medical domain~\cite{karimi2020deep}. 
Our study offers a comprehensive review of the recent advancements in this field. Second, instead of focusing only on modeling techniques to handle label noise, our approach adopts a problem-centric perspective and takes a deeper look into the entire process of noisy label management, including noise detection, noise handling, and evaluation. Typical scenarios where label noise may happen are also audited. Through our survey, we hope to draw greater attention to this underexplored challenge and suggest potential avenues for future research.
%

\section{Methods}
\label{search_methods}

Our scoping review follows the Preferred Reporting Items for Systematic Reviews and Meta-Analyses (PRISMA) guidelines~\cite{tricco2018prisma}.

\subsection{Eligibility criteria}

We overview research on noisy label detection and handling in disease diagnosis and prognosis using deep learning methods. Segmentation problems are excluded from the scope of this study. The inclusion criteria for our review consisted of English-language articles published between 2016 and 2023, including both conference papers and journal articles. We chose this time frame to capture the most up-to-date research in this rapidly evolving field. Additionally, we refer to relevant preprint articles to ensure we consider cutting-edge research that has yet to be published in peer-reviewed venues.

\subsection{Information sources}
A search of multiple databases was carried out, including PubMed (\url{https://pubmed.ncbi.nlm.nih.gov}), the Institute of Electrical and Electronics Engineers (IEEE) Xplore Digital Library (\url{https://ieeexplore.ieee.org}), Google Scholar (\url{https://scholar.google.com}), and Semantic Scholar (\url{https://www.semanticscholar.org}). Important references are also tracked accordingly. The most recent search was executed on Dec 15, 2023. 

\subsection{Search strategy}

All the studies collected in this research were confined to the medical field. Our search terms include ``noisy label AND medical'', ``uncertainty AND medical'', and ``noise AND medical''.

\subsection{Study selection}

Each article was reviewed and summarized by the source of label noise, problem space (e.g., computer vision and Natural Language Processing (NLP)), medical problem (e.g., skin cancer and chest image), label noise detection method, label noise handling method, whether a clean dataset is needed, backbone model, evaluation method and publishing year. During the screening and the full-text review stages, we excluded review articles, non-medical articles, image segmentation-related problems, data denoising, and feature denoising problems.

\section{Results}

\subsection{Included studies and datasets}
\subsubsection{Summary of included literature}
\begin{figure}[h]
\centering
  \includegraphics[width=.5\textwidth]{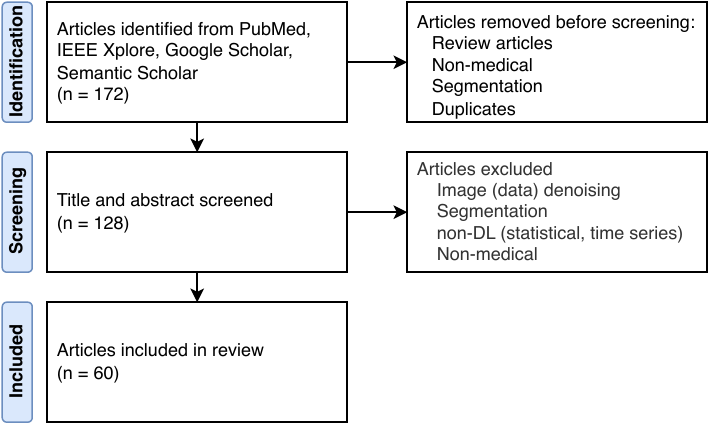}
  \caption{Flowchart of the search strategy.}\label{fig:search_flow}
\end{figure}

A total of 172 articles were retrieved from the 4 databases, of which 60 met the inclusion criteria for our review (Figure~\ref{fig:search_flow}). Additionally, 21 supplementary articles partially addressed the problems of interest and were selectively referenced in the relevant sections. Details and a summary of the included papers are listed in Table~\ref{tab:summary_all_papers}.

{
\vspace{1em}
\tiny\tabcolsep=2pt
\definecolor{lavender}{rgb}{0.9, 0.9, 0.98}

\rowcolors{2}{}{lavender}

\begin{longtable}{c>{\raggedright}p{4em}>{\raggedright}p{1.5cm}>{\raggedright}p{1.5cm}>{\raggedright}p{1.5cm}>{\raggedright}p{1cm}>{\raggedright}p{1.5cm}>{\centering}p{1cm}>{\centering}p{1cm}>{\centering}p{1cm}>
{\centering}p{1cm}>{\raggedright}p{2cm}>{\centering\arraybackslash}p{1cm}}
\caption{Summary of all articles. Clean - the training dataset is noisy and the evaluation dataset is verified to be high quality. Inject - noise injected into training data. Same - training and testing are done on the same dataset. Inherent - the underlying problem is very hard to label. Disagreement - disagreement among experts or inter-expert variability. Curriculum - curriculum learning. OCT - Optical coherence tomography. WSI-Whole slide images} \label{tab:summary_all_papers}\\
\toprule
Ref. & ML scope & Modality & Disease & Noise source & Detection method & Handling method & Evaluation method & Need clean data & Backbone NN& Year & Dataset (size) & Is dataset public \\ \midrule
\endfirsthead

\multicolumn{13}{l}%
{{\bfseries \tablename\ \thetable{} -- continued from previous page}} \\
\toprule
Ref. & ML scope & Modality & Disease & Noise source & Detection method & Handling method & Evaluation method & Need clean data & Backbone NN& Year & Dataset (size) & Is dataset public \\ \midrule
\endhead

\rowcolor{white}
\midrule 
\rowcolor{white}\multicolumn{13}{r}{Continued on next page} \\
\endfoot

\bottomrule
\endlastfoot

\cite{deng2016deep} & CV & MRI & Brain disease & Inherent & Loss & Reweight, robust loss & Inject & N & MLP&2016&Yale-B (2,048), COIL-100 (7,200), Scene-15 (4,485)
 & Y\\
\cite{dgani2018training} & CV& X-ray & Mammography & Inherent & -- & Noise adaption layer & Inject & N & MLP&2018&DDMS (55,890)&Y \\
\cite{xue2019robust} & CV&Dermoscopy & Melanoma & Inherent & Loss & Reweight, curriculum & Inject & N & ResNet&2,019&ISIC-2019 (3,582)& Y\\
\cite{ju2022improving} & CV &Dermoscopy, WSI, fundus photographs & Skin lesion, prostate cancer, retinal disease & Inherent,  disagreement & Disagreement, uncertainty & Reweight, filter, curriculum & Clean & Y & ResNet&2021&ISIC-2019 (3,582), Gleason 2019 (331), Retinal disease Kaggle (88,702 )& Y\\
\cite{xue2022robust} & CV&Dermoscopy, WSI, X-ray & Melanoma, lymph node, lung disease, prostate cancer & Inherent & Co-training & Contrastive loss & Inject, clean & N & ResNet&2022&ISIC-2017 (2,600), ISIC archive (1,582), Kaggle histopathologic (220,025), Gleason 2019 (331)&Y\\
\cite{jiang2023label} & CV&OCT, WSI& Retinal disease, blood cell, colon pathology & Inherent & -- & Contrastive loss, mixup & Inject & N & ResNet&2023&OCT (2,100), Blood cell (4,200), Colon pathology (10,250)&Y\\
\cite{hu2021deep} & CV&CT & Covid & Inherent & -- & Robust loss & Same & N & VGG&2021&6982 &N \\
\cite{lee2020fully} & CV&OCT & Fbrocalcifc plaques  & Disagreement & F1 score & Active learning, label refinement & -- & N & CNN&2020&6,556&Need Request \\
\cite{kurita2023accurate} & CV& WSI & Cervical cancer & Pseudo label & Known & Label refinement, regularization & Clean & Y & Efficient Net& 2023&56,996&N\\ 
\cite{fries2019weakly} & CV&MRI & Aortic valve malformations  & Pseudo label & Explicit & Robust noise & Clean & N & CNN-LSTM& 2019&UKbiobank (4,000)& Need request\\
\cite{wong2022deep} & CV&WSI & AD & Inherent, disagreement & Disagreement & Label refinement & Clean & N & CNN&2022&29 WSI (153,050)&Y \\
\cite{ashraf2022loss} & CV&WSI & Stomach cancer & Inherent & Loss & Label removal & Inject & N & Dense Net& 2022&905 WSI (139,650)&Y\\
\cite{ying2023covid} & CV&X-ray & Covid & Inherent & -- & Label refinement, weighting & Inject & N & CNN&2023&Tawsifur (931), Skytells (1,071), CRs (818) & Y \\
\cite{lopez2021learning} & CV& WSI & Tumor, stroma, and immune infiltrates & Crowdsourcing & -- & Label refinement & Clean & N & Gaussian process&2021&151 WSI &N \\
\cite{karimi2020learning} & CV& MRI& Brain lesion & Inherent & Uncertainty & Label refinement & Clean & N & CNN&2020&165&N \\
\cite{dikici2022advancing} & CV&MRI & Brain metastases & Pseudo label & Known & Label refinement & Clean & Y & Cropnet&2022&217&N \\
\cite{momeny2021learning} & CV &Xray& Lung disease & Inherent & -- & Label refinement & -- & N & AlexNet, ResNet&2021&1,248&Y \\
\cite{jaiswal2022ros} & CV &Dermoscopy, X-ray& Skin lesion, lung disease & Inherent, disagreement, machine label & -- & Robustness & -- & N & ResNet&2022&Skin (25,331), NIH Chest X-ray (112,120)&Y\\
\cite{pulido2020semi} & CV & WSI&Histology & Pseudo label & Known & Model robustness & Clean & Y & ResNet&2022&387&N \\
\cite{paul2021generalized} & CV & X-ray & Lung disease & Machine label & Known & Label refinement & Clean & N & MVSE& 2021&NIH-900 (11,014), Open-i (508,531), PMC (103), CheXpert (4,236)& Y\\
\cite{li2021pathal} & CV & WSI &Pathology & Pseudo label & Loss & Label refinement ,  curriculum & Clean & Y & Efficient Net&2022&Kaggle PANDA (11,000)&Y \\
\cite{gu2018reliable} & CV &Endomicroscopy &Gastrointestinal disease & Pseudo label & Uncertainty & Active learning, label refinement & Inject & Y & ResNet&2019&CLE (1,366)&Y \\
\cite{zhang2021reur} & CV &WSI &Pathology & Pseudo label & Loss & Label refinement, reweight & Inject & N & Large NN& 2021&MICCAI DigestPath2019 (455)&Y\\
\cite{gundel2021robust} & CV &X-ray& Lung disease & Machine label & Model performance & Reweight & Clean & Y & DenseNet&2021&29,7541&Y\\
\cite{algan2020deep} & CV &Retina image & Retinopathy of Prematurity & Disagreement & Meta-learning & Label refinement & Clean & Y & ResNet&2021&1,947& N\\
\cite{ghesu2019quantifying} & CV &X-ray & Lung disease & Inherent & Uncertainty & Reweight, label filter & Clean & N & DenseNet& 2019&NIH Chest X-ray (112,120), PLCO (185,421)&Y\\  
\cite{pham2021interpreting} & CV &X-ray & Lung disease & Explicit & -- & Label smoothing & Clean & Y & DenseNet&2021&CheXpert (224,316)& Y\\
\cite{irvin2019chexpert} & CV&X-ray & Lung disease & Explicit & -- & Label filter, label refinement, model as separate class & Clean & N & DenseNet, ResNet&2019&CheXpert (224,316)&Y \\
\cite{li2023learning} & CV&Dermoscopy,WSI & Skin lesion, histopathology & Disagreement & -- & Label filter & Inject & N & ResNet&2023&Skin (10,000), histopathology (3,000)&Y \\
\cite{chen2022adaptive} & CV&CT & Otosclerosis & Inherent, disagreement & -- & Robust loss & Clean & N & W-Net&2022&902&N \\
\cite{xiang2023automatic} & CV&WSI & Prostate cancer & Disagreement & Loss & Filter label & Clean & N & ResNet&2023 &PANDA (110,000)&Y\\
\cite{del2023labeling} & CV&X-ray & Lung disease & Expertise & Explicit & Robust architecture & Clean & N & VGG& 2023&271&N\\
\cite{calli2019handling} & CV&X-ray & Lung disease& Noisy dataset & Loss, uncertainty & - & Inject, clean & N & ResNet&2019& NIH Chest X-ray (112,120)& Y\\ 
\cite{braun2022influence} & CV& MRI& Lumbar spine stenosis & Disagreement & Influence score & Reweight & Clean & N & MLP& 2022&26,728&N\\
\cite{jimenez2022curriculum} & CV& X-ray& Proximal femur fractures & Expertise, disagreement & Known, uncertainty & Reweight, curriculum & Same & N & CNN& 2022&780&N\\
\cite{liu2021co} & CV&Dermoscopy, WSI & Skin cancer, lymph node & Crowdsourcing & Loss & Label refinement, curriculum, gradient based, label update & Inject & N & ResNet&2021&ISIC-Archive (23,906), PatchCamelyon (327,680)&Y \\
\cite{hu2023fundus} & CV& Fundus image& Fundus disease & Inherent & Loss & Label filter, robust loss & Inject & N & EfficientNet&2023&private data (8),999, ODIR (9,144)&Y/N \\
\cite{shi2021correcting} & CV& OCT& Vulnerable plaque & Pseudo label & Known & Gradient based, label update & Clean & N & ResNet&2021&IVOCT-2017 (3,000)&Y \\
\cite{gao2023clinical} & CV&WSI & Thyroid nodule & Disagreement & Model performance & Reweight & Clean & N & Small CNN&2023&5232&N \\
\cite{gao2022bayesian} & CV&OCT & Retina disease & Inherent & Loss & Label refinement & Inject & N & VGG, inception, ResNet&2022&OCT (4,000), Messidor (1,187), ANIMAL-10N (55,000)&Y\\
\cite{zhou2023refixmatch} & CV& Dermoscopy& Skin lesion & Pseudo label & Explicit & Label smoothing, regularization & Clean & Y & DenseNet&2023&ISIC-2018 (10,015), ISIC-2019 (25,531)&Y \\
\cite{zhu2023robust} & CV&Dermoscopy & Skin cancer & Disagreement & Loss & Label refinement & Clean & N & ResNet& 2023&&\\
\cite{javadi2022training} & CV&Ultra sound & Prostate cancer & Inherent & Uncertainty & Label refinement & Same & N & DNN&2022&353&N\\
\cite{chen2023combating} & CV&Dermoscopy & Skin cancer & Pseudo label & Loss & Robust loss, mixup & Inject & N & ResNet& 2023&24,691&Y\\
\cite{chen2023bomd} & CV&X-ray & Lung disease & Machine label & Cluster based & Label refinement & Inject & N & DenseNet&2023&NIH Chest X-ray14 (112,120), CheXpert (224,316), PadChest (160,861), NIH-Google (1,962)& Y\\

\cite{boughorbel2018alternating} & NLP&EHR & Preterm birth & De-identification & Known & Loss correction & Clean & Y & LSTM&2018&23,578&N \\
\cite{yang2023addressing} & NLP& EHR& Covid & Data entry error, system error, diagnose error & Known & Robust loss, label smoothing & Inject & N & MLP&2023&37,996&Y \\
\cite{murray2019automated} & NLP& Medical data, clinical notes & Lupus erythematosus & Machine label & Explicit & Improve robustness & Clean & N & LR&2019&Chest X-ray14 (112,120), CheXpert (224,316), OpenI (7,470), PadChest (160,861)&Y\\
\cite{dhrangadhariya2023not} & NLP& EHR& PICO entity extraction & Machine label & Explicit & Robust loss & Clean & N & PubMedBERT&2023&23,172&N\\
\cite{li2021semi} & NLP& EHR & Mix of disease & Disagreement & Others & Adversarial training & Same & N & BiLSTM&2021&2,500&Y\\

\cite{vazquez2022label} & Serial&ECG & Cardiac abnormalities & Experience level & Cluster based & Label correction & Clean & N & 1-d CNN&2022&CURIAL (37,896)&Need request\\
\cite{de2023stochastic} & Serial&ECG & Cardiac abnormalities & Disagreement & Co-training & Co-training & Clean & N & Small CNN&2023&UCSF EHR (499,855)&N \\
\cite{baghel2020automatic} & Serial& PCG& Cardiac diseases & Inherent & -- & Label refinement & Inject & N & CNN&2020&PICO (4,081)&Y \\
\cite{vazquez2021two} & Serial&ECG &Cardiac diseases  & Machine label & Cluster based & Label refinement & -- & N & 1-d CNN& 2021&CCKS 2020 (2,050)&Y\\
\cite{Ding2022LearningFA} & Serial&PPG & Atrial Fibrillation & Pseudo label & Cluster based & Robust loss & Clean & Y & ResNet& 2022&CINC 2021 (10,1149)&Y\\
\cite{hong2023semi} & Survival&EHR & Lung cancer, covid & Pseudo label & Explicit & Label refinement & Clean & Y & ML&2023&PTB-XL (21,837), MRI Sunnybrook challenge (153)&Y \\
\cite{ren2023ocrfinder} & Sequential  model&Genome & Chromatin region detection & Pseudo label & Loss function, uncertainty & Robust loss & Clean & N & CNN-LSTM&2023&1,000& Y\\
\cite{tjandra2023leveraging} & ML& Regular data& Acute respiratory failure, shock in ICU & Inherent & NN to estimate, transition prob & Reweight & Inject & Y & MLP&2023&CINC 2021 (101,149)&Y \\
\cite{vernekar2022improving} & ML& Regular data& Diabete & Inherent & -- & Label refinement & Clean & Y & MLP&2022&24,100& N\\
\cite{xu2020hybrid} & ML&Regular data & Mix of disease & Inherent & Cluster based & Label filtering & Inject & N & ML&2020&Mix of datasets, each $\le$ 1,000&Y\\
\end{longtable}}
\subsubsection{Overview of problem space}
Most of the prediction problems in the included studies are disease diagnosis problems. Due to the wide availability of data, Chest X-ray and skin cancer are among the top medical domains that have been extensively explored. Some medical specialties are hard to diagnose by nature, which in turn causes the prevalence of label noise. For instance, Xue et al. demonstrate images indicating that the malignant histology lymph node images present quite similar colors and structures to benign ones \cite{xue2022robust}. The same situation can be observed for skin lesion images as well. 

Table~\ref{tab:summary_all_papers} lists “disease” and “noise source” for detailed information. Since most of the problems are diagnosis problems, “labels” are usually the disease of interest. The ``Noise source" column explains the reason that causes label noise. ``Modality", ``ML scope," and ``Backbone NN" columns summarize the problem space from a deep learning perspective. The selection of the backbone model usually depends on the domain and data size. Since most problems are image-based, ResNet and DenseNet are widely adopted. If the dataset size is small, customized small CNN can be an alternative. The rest of the columns will be elaborated in the following sections.
\subsection{Source, impact, and detection of label noise}
\label{source_impact_detection}

\subsubsection{Sources of label noise}
\label{source_of_label_noise}

The primary source of label noise originates from the inherent complexity of the medical problem being examined. Multiple factors such as annotators' fatigue or distraction~\cite{brady2017error,lu2019automated} or varying levels of expertise among annotators may contribute to label noise~\cite{lu2019automated, ashraf2022loss, jaiswal2022ros, chen2022adaptive, del2023labeling, jimenez2022curriculum}. 
Furthermore, there are inherent challenges in specific medical domains, such as the similarity between skin cancer and benign conditions~\cite{li2023learning, ju2022improving, xue2022robust, jaiswal2022ros, xue2019robust, hekler2020effects}. 
Such complexity of medical diagnosis is usually reflected by intra-observer and inter-observer variability, seen across multiple areas from Alzheimer's Disease (AD) to retinopathy~\cite{lee2020fully, ju2022improving, algan2020deep, vazquez2022label, wong2022deep, jaiswal2022ros, chen2022adaptive, de2023stochastic, li2023learning, xiang2023automatic, gao2023clinical, zhu2023robust}.\
Some studies have attempted to explicitly quantify the proportion of intra- and inter-observer variability, finding a range of 5\% to 40\%, depending on the specific specialties and datasets~\cite{campbell2017plus, wallace2008agreement,li2023learning,chen2022adaptive,braun2022influence,lee2020fully}. 

Another significant contributor to label noise is the generation of labels by non-experts or non-humans. These labels may be extracted from an automated system~\cite{murray2019automated, dhrangadhariya2023not}, developed as pseudo labels in semi-supervised learning~\cite{ren2023ocrfinder, fries2019weakly}, stemmed from crowdsourcing~\cite{lopez2021learning, liu2021co}, or be self-reported or non-verified~\cite{hekler2020effects, cosentino2023inference, ding2022impact}.
It is worth noting that although crowdsourcing is proven to be highly influential within the traditional computer vision (CV) community, its efficacy needs to improve in medical research due to the specialized expertise required \cite{lopez2021learning}.

Besides the above-mentioned two types of contributors, label noise can also be attributed to other factors such as the de-identification process \cite{boughorbel2018alternating}, data entry errors \cite{xue2019robust, pechenizkiy2006class}, incomplete information \cite{xue2019robust}, as well as data acquisition, transmission, and storage errors \cite{baghel2020automatic, vazquez2021two, ying2023covid}. Notably, some popular datasets, like ChestX-Ray14, are widely known to be noisy \cite{ghesu2019quantifying, irvin2019chexpert}.

\subsubsection{Impact of label noise}

The impact of label noise is reflected by its detriment to classification performance, which is usually measured by accuracy, AUC, PRAUC, and F1. For multi-class prediction problems, the macro average of metrics is usually used. Many studies aim to evaluate the impact of label noise on the performance of deep learning models. While some studies argued that deep learning is resilient to corrupted labels \cite{potapenko2022detection}, they failed to assess the scenario where noisy labels were properly handled, which could potentially yield even better outcomes. In contrast, most studies have demonstrated a substantial detrimental effect when label noise is not addressed appropriately \cite{zhu2023robust, hekler2020effects, ding2022impact, khanal2023investigating, samala2020generalization, buttner2023impact, jang2020assessment}.
For example, research by Samala et al. highlights that as little as 10\% of injected label noise can greatly increase DNN's generalization error \cite{samala2020generalization}.
Jang et al. found that even a 2\% random flip of labels significantly affects chest X-ray prediction accuracy \cite{jang2020assessment}.
Collectively, these findings suggest that the impact of label noise is indeed remarkable, particularly in cases with relatively small sample sizes.

Apart from its effect on prediction accuracy, label noise could also influence fairness, especially when the noise varies across different patient subgroups \cite{petersen2023path, tjandra2023leveraging}.
For instance, Tjandra et al. pointed out that females are more frequently underdiagnosed for cardiovascular diseases \cite{tjandra2023leveraging}.

\subsubsection{Detection of noisy labels}
\label{sec:detection_label_noise}

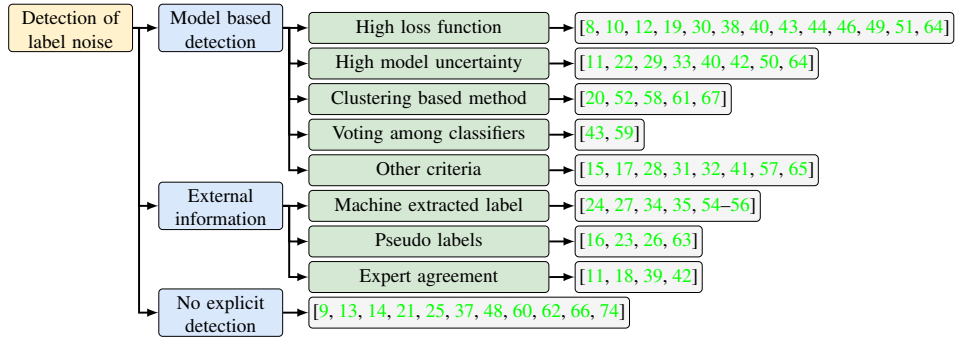
\begin{figure}[ht]
\vspace{1em}
\centering
\resizebox{.8\linewidth}{!}{
\definecolor{myblue}{HTML}{DAE8FC}
\definecolor{myyellow}{HTML}{FFF2CC}
\definecolor{mygreen}{HTML}{D5E8D4}
\definecolor{myred}{HTML}{F8CECC}
\definecolor{mygrey}{HTML}{F5F5F5}

\tikzset{
root/.style  = {
    draw, 
    text width=60pt, 
    rounded corners=2pt,
    align=center, 
    anchor=west, 
    rectangle, 
    font=\small, 
    inner ysep=2pt, 
    inner xsep=2pt,
    fill=myyellow,
},
first/.style  = {
    draw, 
    text width=60pt, 
    rounded corners=2pt,
    align=center, 
    anchor=west, 
    rectangle, 
    font=\small, 
    inner ysep=2pt, 
    inner xsep=2pt,
    fill=myblue,
},
second/.style  = {
    draw, 
    text width=120pt, 
    rounded corners=2pt,
    align=center, 
    anchor=west, 
    rectangle, 
    font=\small, 
    inner ysep=2pt, 
    inner xsep=2pt,
    fill=mygreen,
},
leaf/.style  = {
    draw, 
    rounded corners=2pt,
    align=left, 
    anchor=west, 
    rectangle, 
    font=\small, 
    inner ysep=2pt, 
    inner xsep=2pt,
    fill=mygrey,
},
}

\begin{forest} 
for tree={
    calign=child edge,
    calign child=1,
    grow'=east,
    growth parent anchor=west,
    parent anchor=east,
    child anchor=west,
    edge={line width=1pt},
    edge path={\noexpand\path[\forestoption{edge},<-, >={latex}] 
          (.child anchor)  -- +(-10pt,0pt) |- (!u.parent anchor)
         \forestoption{edge label};},
    s sep=2pt,
},
before typesetting nodes={
    for tree={
        content/.wrap value={\strut #1},
    }
}
[Detection of label noise, root,
    [Model based detection, first,
        [High loss function, second,
            [\cite{gao2023clinical, xue2019robust, ashraf2022loss, xue2022robust, ren2023ocrfinder, zhang2021reur, deng2016deep, zhu2023robust, xiang2023automatic, liu2021co, calli2019handling, hu2023fundus,chen2023combating}, leaf]
        ]
        [High model uncertainty, second,
            [\cite{ju2022improving, jimenez2022curriculum, ren2023ocrfinder, gu2018reliable, ghesu2019quantifying, calli2019handling, karimi2020learning,javadi2022training}, leaf]
        ]
        [Clustering based method, second,
            [\cite{vazquez2022label, ying2023covid, vazquez2021two,chen2023bomd,xu2020hybrid}, leaf]
        ]
        [Voting among classifiers, second,
            [\cite{de2023stochastic, liu2021co}, leaf]
        ]
        [Other criteria, second,
            [\cite{lee2020fully, tjandra2023leveraging, algan2020deep, braun2022influence, gundel2021robust, fries2019weakly, li2021pathal,li2021semi}, leaf]
        ]
    ]
    [External information, first,
        [Machine extracted label, second,
            [\cite{murray2019automated, dhrangadhariya2023not, paul2021generalized, irvin2019chexpert, pham2021interpreting, yang2023addressing,momeny2021learning}, leaf]
        ]
        [Pseudo labels, second,
            [\cite{kurita2023accurate,hong2023semi,dikici2022advancing,pulido2020semi}, leaf]
        ]
        [Expert agreement, second,
            [\cite{del2023labeling,ju2022improving, jimenez2022curriculum, wong2022deep}, leaf]
        ]
    ]
    [No explicit detection, first,
        [\cite{jiang2023label, jaiswal2022ros, chen2022adaptive, hu2021deep, Ding2022LearningFA, zhou2023refixmatch, lopez2021learning, pechenizkiy2006class, baghel2020automatic, vernekar2022improving, dgani2018training}, leaf]
    ]
]
\end{forest}
}
\caption{Summary of noisy label detection methods. Terminologies in the figure are explained in the glossary.}\label{fig:detection}
\end{figure}

Many strategies adopt a two-step or iterative approach: label noise detection and handling. In this section, we summarize the detection methods (Figure \ref{fig:detection}). Not all label noise requires detection; some are evident by nature. In semi-supervised learning, pseudo-labels are known to be noisy. In supervised learning, expert disagreements on disease diagnosis also indicate label noise, which can be quantified by Cohen's Kappa coefficient \cite{del2023labeling, ju2022improving, wong2022deep}.

When the degree of label noisiness is unknown, machine learning based detection methods can be applied. The most widely used criteria are the first-order information (e.g., prediction magnitude or loss function) and the second-order information (model or epistemic uncertainty). The study by Calli et al. shows that they are both valid in assessing label noisiness \cite{calli2019handling}. The difficulty of using them is the selection of threshold when the noise rate is still being determined \cite{liu2021co, ren2023ocrfinder, xiang2023automatic, zhu2023robust, liu2015classification}.
Despite the popularity of these two criteria, other studies indicate that a high loss function could also reflect a correct but hard-to-learn instance. Therefore, they try to disentangle the effects of noisy labels, hard instances, and class imbalance \cite{li2021pathal, li2023learning}.

Besides the loss function and model uncertainty, other machine-learning approaches are used to detect noisy labels. Ying et al. have used the clustering-based method to identify prototypes or clusters within the feature space and assume labels in the same cluster are similar \cite{ying2023covid}. The same strategy can also be used in label refinement (Section \ref{sec:refinement_filtering}). Braun et al. have developed another method to detect noise by evaluating the self-influence of a sample. This is defined by the change in the test sample loss function due to the inclusion or exclusion of the candidate instance to the training set \cite{braun2022influence}.
The underlying logic is that a noisy sample can decrease its own training error, but not the error of a test sample. Algan et al. have employed a meta-learning criterion to search for optimal soft-labels that contribute to the most noise-robust training \cite{algan2020deep}.

\subsection{Deep learning methodologies}
\label{methods}

\begin{figure}[ht]
\vspace{1em}
\centering
\resizebox{.8\linewidth}{!}{
\definecolor{myblue}{HTML}{DAE8FC}
\definecolor{myyellow}{HTML}{FFF2CC}
\definecolor{mygreen}{HTML}{D5E8D4}
\definecolor{myred}{HTML}{F8CECC}
\definecolor{mygrey}{HTML}{F5F5F5}

\tikzset{
root/.style  = {
    draw, 
    text width=80pt, 
    rounded corners=2pt,
    align=center, 
    anchor=west, 
    rectangle, 
    font=\small, 
    inner ysep=2pt, 
    inner xsep=2pt,
    fill=myyellow,
},
first/.style  = {
    draw, 
    text width=80pt, 
    rounded corners=2pt,
    align=center, 
    anchor=west, 
    rectangle, 
    font=\small, 
    inner ysep=2pt, 
    inner xsep=2pt,
    fill=myblue,
},
second/.style  = {
    draw, 
    text width=120pt, 
    rounded corners=2pt,
    align=center, 
    anchor=west, 
    rectangle, 
    font=\small, 
    inner ysep=2pt, 
    inner xsep=2pt,
    fill=mygreen,
},
leaf/.style  = {
    draw, 
    rounded corners=2pt,
    align=left, 
    anchor=west, 
    rectangle, 
    font=\small, 
    inner ysep=2pt, 
    inner xsep=2pt,
    fill=mygrey,
},
}

\begin{forest} 
for tree={
    calign=child edge,
    calign child=1,
    grow'=east,
    growth parent anchor=west,
    parent anchor=east,
    child anchor=west,
    edge={line width=1pt},
    edge path={\noexpand\path[\forestoption{edge},<-, >={latex}] 
          (.child anchor)  -- +(-10pt,0pt) |- (!u.parent anchor)
         \forestoption{edge label};},
    s sep=2pt,
},
before typesetting nodes={
    for tree={
        content/.wrap value={\strut #1},
    }
}
[Methods to handle label noise, root,
    [Label refinement and filtering, first,
        [Co-teaching, second,
            [\cite{liu2021co, de2023stochastic, zhu2023robust, ren2023ocrfinder,javadi2022training}, leaf]
        ]
        [Teacher-student, second,
            [\cite{kurita2023accurate}, leaf]
        ]
        [Label correction module, second,
            [\cite{zhang2021reur}, leaf]
        ]
        [Active learning, second,
            [\cite{lee2020fully, lu2019automated, gu2018reliable}, leaf]
        ]
        [Cluster based, second
            [\cite{vazquez2022label, ying2023covid, vazquez2021two, vernekar2022improving,chen2023bomd}, leaf]
        ]
        [Gradient based, second,
            [\cite{shi2021correcting,liu2021co}, leaf]
        ]
        [Other label refinement, second,
            [\cite{algan2020deep, paul2021generalized, lopez2021learning}, leaf]
        ]
        [Removing noisy labels, second,
            [\cite{li2023learning, ju2022improving, ashraf2022loss, xiang2023automatic, li2021pathal, ghesu2019quantifying,xu2020hybrid}, leaf]
        ]
    ]
    [Loss and architecture change, first,
        [Architecture change, second,
            [\cite{zhang2021reur, dgani2018training}, leaf]
        ]
        [Loss correction, second,
            [\cite{boughorbel2018alternating, tjandra2023leveraging}, leaf]
        ]
    ]
    [Loss and model robustness improvement, first,
        [Noise robust loss function, second,
            [\cite{chen2022adaptive, hu2021deep, deng2016deep, Ding2022LearningFA, ren2023ocrfinder, yang2023addressing, hu2023fundus,chen2023combating}, leaf]
        ]
        [Mixup / mixmatch, second,
            [\cite{yang2023addressing, pham2021interpreting, pulido2020semi, jiang2023label, li2023learning,chen2023combating}, leaf]
        ]
        [Label smoothing, second,
            [\cite{zhou2023refixmatch, pham2021interpreting, yang2023addressing}, leaf]
        ]
        [Contrastive learning, second,
            [\cite{jiang2023label, xue2022robust}, leaf]
        ]
    ]
    [Training paradigm change, first,
        [Weighting, second,
            [\cite{ju2022improving, xue2019robust, deng2016deep, braun2022influence, murray2019automated, dhrangadhariya2023not, fries2019weakly, zhang2021reur, ghesu2019quantifying}, leaf]
        ]
        [Curriculum learning, second,
            [\cite{ju2022improving, xue2019robust, li2021pathal, liu2021co}, leaf]
        ]
    ]
    [Others, first,
        [\cite{irvin2019chexpert, pechenizkiy2006class, momeny2021learning, wong2022deep, algan2020deep, jaiswal2022ros, del2023labeling, lee2020fully,li2021semi}, leaf]
    ]
]
\end{forest}
}
\caption{Methods applied to handle label noise. Terminologies in the figure are explained in the glossary.} \label{fig:method_summary}
\end{figure}
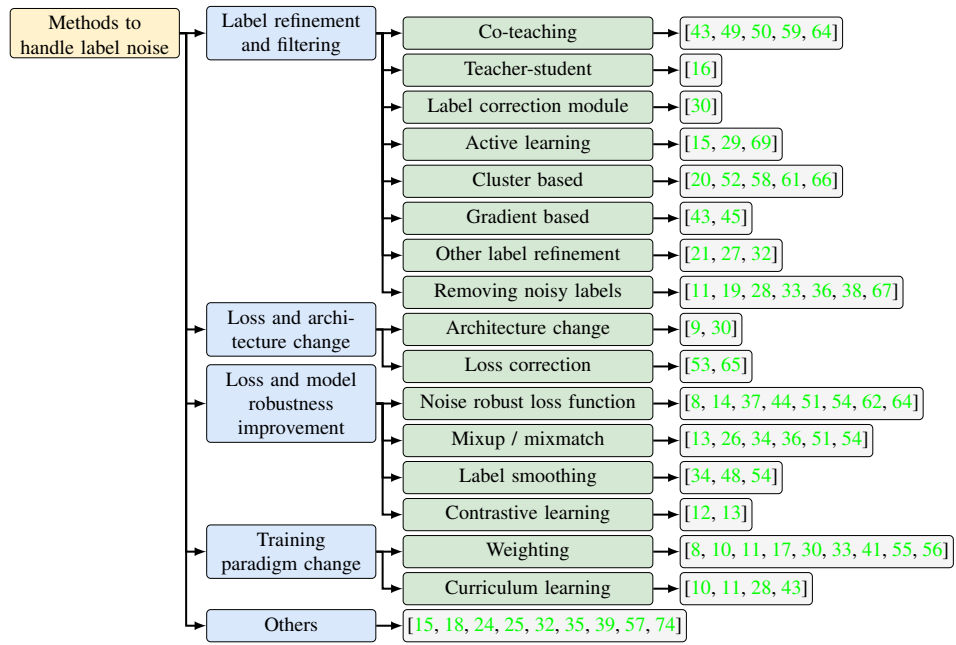

Techniques for addressing noisy labels in medical research can be organized into four main categories (Figure~\ref{fig:method_summary}). Some studies only utilize one technique, while others leverage multiple techniques in the same study. Further categorization of these methods can be done based on whether they require a clean dataset for training or if they are applied iteratively.

\subsubsection{Label filtering and refinement}
\label{sec:refinement_filtering}

The label filtering process starts with applying noise detection methods (Section~\ref{sec:detection_label_noise}), followed by a straightforward approach to discarding noisy instances. One disadvantage of discarding noisy instances is the potential removal of genuinely complex but valuable data. Hence, the weighting method (Section~\ref{sec:training_paradigm}) or label refinement are better alternatives \cite{braun2022influence}.

Label refinement is another approach that takes a more conservative standpoint by revising noisy labels rather than discarding them entirely. One way is through the use of a dedicated classifier \cite{vernekar2022improving}.
Another approach involves utilizing the stochastic gradient descent to update labels \cite{liu2021co, shi2021correcting}, where the gradient is calculated with respect to labels. After the label refinement step, original labels may still be used in model training \cite{vazquez2022label}.

\subsubsection{Neural network architecture change}

Architecture change explicitly models the transition between latent ground true labels and observed noisy labels through dedicated layers. However, it is less common in medical research due to the constraint of a limited sample size.  The most common changes to architecture involve employing loss correction or adaptation layers.

Loss correction explicitly estimates the transition matrix between the observed label and the latent ground truth label, denoted as $T$ ($T_{i,j}=P(\Tilde{y}=e^j|y=e^i)$) \cite{patrini2017making}, where $\Tilde{y}$ and $y$ are the noisy and true labels.
The estimated transition matrix is then used to weight the loss function $l_{corrected}=T^{-1}l\left(\hat{p}(y|x)\right)$. It is proved that the corrected loss function is unbiased.

Differently, the noisy adaptation layer integrates this noise transition process directly into the neural network by adding an extra softmax layer after the original one \cite{goldberger2016training}.
However, this additional layer makes the neural network non-identifiable, which poses challenges to optimization, and a careful initialization is needed.

\subsubsection{Enhancing model robustness}

Instead of directly addressing noisy labels, an alternative workstream focuses on enhancing model robustness and reducing overfitting.

Designing new loss functions is the most popular approach to improve robustness. The issue with the cross-entropy loss in regular classification applications is that its gradient can sharply increase as the predicted probability approaches the opposite of the corresponding label. While this accelerates convergence, it leads to overfitting when the label is wrong. Loss functions that suppress gradient can mitigate this issue. Interestingly, mean square error, the default loss function in regression, has demonstrated resilience to label noise in classification problems \cite{algan2020deep, ren2023ocrfinder}.
Another approach to construct robust loss is to apply similar motivations as contrastive learning, which enforces instances close in latent feature space to share similar labels \cite{deng2016deep, Ding2022LearningFA, yang2023addressing}.

In addition to robust loss functions, regularization methods such as label smoothing and mix-up can also be employed. Regularization can also be achieved by customizing the optimization strategy. For instance, Hu et al. use sharpness-aware minimization to find a ``flat minimum'' whose neighbors have uniformly low training loss values \cite{hu2023fundus}.

\subsubsection{Training paradigm change}
\label{sec:training_paradigm}

Reweighting and curriculum learning are the most widely used training paradigm changes to tackle label noise. The reweighting strategy adjusts the instance weights by prioritizing clean labels over noisy ones. The concept of curriculum learning, originating from education, starts with training neural networks on easy samples (easy curriculum) and gradually progresses to more complex ones (hard curriculum). It aims to smoothen the warm-up process during optimization and achieve better local minima. In the context of label denoising, training can start with clean labels identified as described in Section \ref{sec:detection_label_noise}.

There are other paradigm changes in addition to curriculum learning and reweighting. Irvin et al. treated noisy labels as a distinct class rather than assigning a pseudo label \cite{irvin2019chexpert}. 
Algan et al. introduced a meta-learning strategy to generate soft labels \cite{algan2020deep}.
The goal is to produce optimal soft labels for noisy instances so that a base classifier trained on such instances would perform best on a separate clean meta-data.

\subsection{Evaluation}

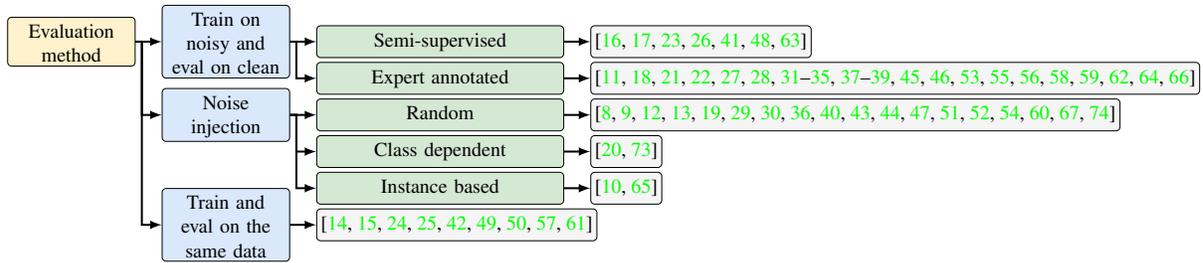
\begin{figure}
\centering
\resizebox{\linewidth}{!}{
\definecolor{myblue}{HTML}{DAE8FC}
\definecolor{myyellow}{HTML}{FFF2CC}
\definecolor{mygreen}{HTML}{D5E8D4}
\definecolor{myred}{HTML}{F8CECC}
\definecolor{mygrey}{HTML}{F5F5F5}

\tikzset{
root/.style  = {
    draw, 
    text width=60pt, 
    rounded corners=2pt,
    align=center, 
    anchor=west, 
    rectangle, 
    font=\small, 
    inner ysep=2pt, 
    inner xsep=2pt,
    fill=myyellow,
},
first/.style  = {
    draw, 
    text width=60pt, 
    rounded corners=2pt,
    align=center, 
    anchor=west, 
    rectangle, 
    font=\small, 
    inner ysep=2pt, 
    inner xsep=2pt,
    fill=myblue,
},
second/.style  = {
    draw, 
    text width=120pt, 
    rounded corners=2pt,
    align=center, 
    anchor=west, 
    rectangle, 
    font=\small, 
    inner ysep=2pt, 
    inner xsep=2pt,
    fill=mygreen,
},
leaf/.style  = {
    draw, 
    rounded corners=2pt,
    align=left, 
    anchor=west, 
    rectangle, 
    font=\small, 
    inner ysep=2pt, 
    inner xsep=2pt,
    fill=mygrey,
},
}

\begin{forest} 
for tree={
    calign=child edge,
    calign child=1,
    grow'=east,
    growth parent anchor=west,
    parent anchor=east,
    child anchor=west,
    edge={line width=1pt},
    edge path={\noexpand\path[\forestoption{edge},<-, >={latex}] 
          (.child anchor)  -- +(-10pt,0pt) |- (!u.parent anchor)
         \forestoption{edge label};},
    s sep=2pt,
},
before typesetting nodes={
    for tree={
        content/.wrap value={\strut #1},
    }
}
[Evaluation method, root,
    [Train on noisy and eval on clean, first,
        [Semi-supervised, second,
            [\cite{braun2022influence, kurita2023accurate, hong2023semi, fries2019weakly, dikici2022advancing, pulido2020semi, zhou2023refixmatch}, leaf]
        ]
        [Expert annotated, second,
            [\cite{li2021pathal, gundel2021robust, Ding2022LearningFA, boughorbel2018alternating, ju2022improving, vazquez2022label, de2023stochastic, algan2020deep, ghesu2019quantifying, pham2021interpreting, irvin2019chexpert, murray2019automated, dhrangadhariya2023not, ren2023ocrfinder, wong2022deep, lopez2021learning, karimi2020learning, paul2021generalized, chen2022adaptive, xiang2023automatic, del2023labeling, shi2021correcting, gao2023clinical, vernekar2022improving}, leaf]
        ]
    ]
    [Noise injection, first,
        [Random, second,
            [\cite{gu2018reliable, dgani2018training, baghel2020automatic, deng2016deep, zhang2021reur, li2023learning, calli2019handling, liu2021co,hu2023fundus, gao2022bayesian, yang2023addressing, xue2022robust, jiang2023label, pechenizkiy2006class, ashraf2022loss,chen2023combating,chen2023bomd,xu2020hybrid}, leaf]
        ]
        [Class dependent, second,
            [\cite{ding2022impact, ying2023covid}, leaf]
        ]
        [Instance based, second,
            [\cite{xue2019robust, tjandra2023leveraging}, leaf]
        ]
    ]
    [Train and eval on the same data, first,
        [\cite{hu2021deep, lee2020fully, momeny2021learning, jaiswal2022ros, vazquez2021two, jimenez2022curriculum, zhu2023robust,javadi2022training,li2021semi}, leaf]
    ]
]
\end{forest}
}
\caption{Evaluation method summary.} \label{fig:eval_method_summary}
\end{figure}

Evaluation is crucial yet challenging when studying noisy label handling, primarily due to the necessity of a clean dataset. Two common evaluation methods were identified from the reviewed studies (Figure~\ref{fig:eval_method_summary}): (1) using data with ground truth for evaluation, and (2) injecting noise into training data (with the original data being used for evaluation).

In the first approach, ground truth data could come from expert annotation, or the labeled part of data in a semi-supervised learning setting \cite{xue2022robust}.
In the second approach where noise is simulated, three different underlying assumptions exist \cite{frenay2014comprehensive}: (1) random noise where the noise probability is independent of both features and latent true labels, (2) class-dependent noise where the noise probability depends only on the class labels, and (3) instance-based noise \cite{Song2023learning}, which assumes the noise mechanism depends on both labels and features.

We suggest the use of ground truth data as an evaluation criterion whenever possible. While instance-based label injection is arguably the most compelling simulation for label noise, we find it is rarely applied in practice. Moreover, a positive ratio should be maintained the same after noise injection to mitigate the confounding impact of class imbalance \cite{yang2023addressing}, and noise may be simulated across various severity \cite{gao2022bayesian}.

%

\section{Discussion}
\label{discussion}

Medical research is intrinsically prone to noisy labels stemming from its inherent ambiguity, variations in label sources relating to data quality, dependence on labels generated by automated systems, and more. A considerable number of the papers we have reviewed report substantial deterioration in model performance due to label noise. In this review, we focus on a comprehensive examination of different aspects related to label noise management.

How label noise is handled depends on factors such as the availability of clean data, sample size, and noise level. Approaches like label refinement and training paradigm optimization, which heavily rely on noise detection, typically require a clean dataset. In contrast, architecture change and robustness improvement usually do not have this requirement. Applying noise-handling techniques iteratively often leads to improved results, with the noise level tending to decrease with each iteration. Due to the uncertainty in noise detection, priority should be given to reweighting or label refinement over filtering out noisy labels \cite{braun2022influence}. To evaluate how well the noisy label issue is addressed, simulations should be conducted using instance-based noise. In real-world applications where addressing noisy labels is not the primary objective, the quality of a clean evaluation dataset becomes crucial.

Several directions may deserve attention for future research. The first is to further distinguish between label noise and genuinely hard examples. A duality can be applied to instances with a high loss function or model uncertainty. They can be perceived either as "noisy instances" in our context or as valuable "hard examples" in hard example mining. In the papers we reviewed, they are deprioritized, whereas in hard example mining literature, they should be emphasized. While some studies have attempted to address this issue, their suggested approaches are often difficult to implement and lack broad applicability. Secondly, learning from noisy labels is closely linked to other domains, such as semi-supervised learning, active learning, uncertainty-aware classification, ensemble learning, and regularization. Cross-disciplinary investigation could provide another promising direction for future research. Thirdly, few studies compare different backbone models' sensitivity and robustness to label noise. This will be helpful in guiding the selection of model architecture. Fourthly, even in the broader deep learning community, there are limited studies comparing different method’s performance and how to more effectively combine them. A systematic review or benchmark study will be helpful. Lastly, label noise detection methods have not been thoroughly discussed in existing literature. Detection methods could not only help to handle noise but also be useful in assessing the quality of evaluation data and estimating the noise ratio.

\section{Conclusion}

In this scoping review, we systematically examined the current state of research on noisy label detection and management in the medical domain. Our findings suggest that the techniques for handling noisy labels in the medical community are currently up to date with the general deep-learning community. Despite their utility, the adoption of label noise management in mainstream medical research remains limited. These methodologies are predominantly discussed in works specifically targeting the issue of noisy labels. We recommend considering label noise as a standard element in medical research. Many of these noisy label-handling methodologies (e.g., reweighting, noise robust loss, curriculum learning) do not require extensive efforts and can seamlessly integrate into existing research workflows.


\section*{Glossary}
\label{glossary}
{
\small\definecolor{lavender}{rgb}{0.9, 0.9, 0.98}

\rowcolors{2}{}{lavender}

\begin{xltabular}{\textwidth}{@{}lX@{}}
\toprule
Active learning & A machine learning algorithm that interactively queries a human (or other information source) to label new data points. Usually, the data points to be queried are selected to maximize the information that can be obtained.\\
Architecture change & Architecture change explicitly models the transition between latent ground true labels and observed noisy
labels through dedicated layers. \\
Cluster-based & Instances are clustered in the feature space. The label of each cluster is usually decided by some ground truth data or majority voting. If the label of an instance disagrees with the cluster label, it might be corrected.\\
Clustering based method & Instances are clustered in the feature space. The label of each cluster is usually decided by some ground truth data or majority voting. Instances whose label is different from their cluster are considered noisy.\\
Co-teaching& Two classifiers are trained with different parts of training data and feature sets. The labels are refined iteratively. In each iteration, each classifier's most confident predictions will be used to refine the label of the training set of the other classifier.\\
Contrastive learning&  A deep learning approach to extract meaningful representations by contrasting positive and negative pairs of instances. It leverages the assumption that similar instances should be closer together in the learned embedding space, while dissimilar instances should be further apart. Contrastive loss is usually more robust to label noise than cross-entropy loss.\\
Expert agreement& Consider a label to be noisy if experts disagree on the annotation.\\
Gradient based& Calculate the gradient with respect to the label instead of parameters and use stochastic gradient descent to update the label. \\
High loss function &  Consider the label to be noisy if the loss function is high.\\
High model uncertainty & Consider the label to be noisy if estimated prediction uncertainty is high.\\
Label correction module& Add a dedicated component in the neural network to refine labels. \\
Label smoothing& A regularization technique that introduces noise to the labels. For a k-class classification problem, labels 0 and 1 are replaced by $\frac{\epsilon}{k-1}$ and 1-$\epsilon$, respectively. Here $\epsilon$ decides the noise level to be introduced. \\
Loss correction& Loss correction explicitly estimates the transition matrix between the observed label and the latent ground truth label, denoted as $T$ ($T_{i,j}=P(\Tilde{y}=e^j|y=e^i)$), where $\Tilde{y}$ and $y$ are the noisy and true labels.
The estimated transition matrix is then used to weight the loss function $l_{corrected}=T^{-1}l\left(\hat{p}(y|x)\right)$.\\
Machine extracted labels & Labels are derived from automated systems such as the NLP model instead of being annotated by experts. It is considered noisy since the automated system is error-prone. \\
Mixup/mixmatch& Mixup is a data augmentation strategy. A new instance is created by taking a weighted average of two existing instances. Here the weighted average is applied to both the features and labels. Mixmatch is a data augmentation strategy built on top of mixup.\\
Pseudo labels& In semi-supervised learning, usually the unlabeled part will be assigned a label by a classifier trained on the labeled part of data. This label is called pseudo label and considered noisy.\\
Removing noisy labels& Remove the instances whose labels are considered noisy.\\
Teacher-student& Use a larger neural network's prediction to refine the label for the training set of a smaller neural network.\\
Voting among classifiers & Consider the label to be noisy if different estimators' predictions disagree\\
\bottomrule
\end{xltabular}
}

\section*{Funding Statement}

This work was supported by NSF CAREER Award No. 2145640 and the National Library of Medicine under Award No. R01LM014306.

\section*{Competing Interests Statement}

There are no competing interests to declare.

\section*{Data Availability statement}

The data underlying this article are available in the article and in its online supplementary material.

\section*{Contributorship Statement}

Study concepts/study design, Y.W., H.J., Y.P.; manuscript drafting or manuscript revision for important intellectual content, all authors; approval of the final version of the submitted manuscript, all authors; agrees to ensure any questions related to the work are appropriately resolved, all authors; literature research, Y.W., Y.D., C.S., M.L.; data interpretation, Y.W., Y.D.; and manuscript editing, all authors.

\bibliographystyle{vancouver}
\bibliography{main}

\begin{thebibliography}{10}

\bibitem{wallace2008agreement}
Wallace DK, Quinn GE, Freedman SF, Chiang MF.
\newblock Agreement among pediatric ophthalmologists in diagnosing plus and pre-plus disease in retinopathy of prematurity.
\newblock Journal of American Association for Pediatric Ophthalmology and Strabismus. 2008;12(4):352-6.

\bibitem{frenay2014comprehensive}
Fr{\'e}nay B, Kab{\'a}n A, et~al.
\newblock A comprehensive introduction to label noise.
\newblock In: ESANN. Citeseer; 2014. .

\bibitem{Song2023learning}
Song H, Kim M, Park D, Shin Y, Lee JG.
\newblock Learning From Noisy Labels With Deep Neural Networks: A Survey.
\newblock IEEE Trans Neural Netw Learn Syst. 2023 Nov;34(11):8135-53.

\bibitem{algan2021image}
Algan G, Ulusoy I.
\newblock Image classification with deep learning in the presence of noisy labels: A survey.
\newblock Knowledge-Based Systems. 2021;215:106771.

\bibitem{liang2022review}
Liang X, Liu X, Yao L.
\newblock Review--a survey of learning from noisy labels.
\newblock ECS Sensors Plus. 2022;1(2):021401.

\bibitem{karimi2020deep}
Karimi D, Dou H, Warfield SK, Gholipour A.
\newblock Deep learning with noisy labels: Exploring techniques and remedies in medical image analysis.
\newblock Medical image analysis. 2020;65:101759.

\bibitem{tricco2018prisma}
Tricco AC, Lillie E, Zarin W, O'Brien KK, Colquhoun H, Levac D, et~al.
\newblock PRISMA extension for scoping reviews (PRISMA-ScR): checklist and explanation.
\newblock Annals of internal medicine. 2018;169(7):467-73.

\bibitem{deng2016deep}
Deng Y, Bao F, Deng X, Wang R, Kong Y, Dai Q.
\newblock Deep and structured robust information theoretic learning for image analysis.
\newblock IEEE Transactions on Image Processing. 2016;25(9):4209-21.

\bibitem{dgani2018training}
Dgani Y, Greenspan H, Goldberger J.
\newblock Training a neural network based on unreliable human annotation of medical images.
\newblock In: 2018 IEEE 15th International symposium on biomedical imaging (ISBI 2018). IEEE; 2018. p. 39-42.

\bibitem{xue2019robust}
Xue C, Dou Q, Shi X, Chen H, Heng PA.
\newblock Robust learning at noisy labeled medical images: Applied to skin lesion classification.
\newblock In: 2019 IEEE 16th International Symposium on Biomedical Imaging (ISBI 2019). IEEE; 2019. p. 1280-3.

\bibitem{ju2022improving}
Ju L, Wang X, Wang L, Mahapatra D, Zhao X, Zhou Q, et~al.
\newblock Improving medical images classification with label noise using dual-uncertainty estimation.
\newblock IEEE transactions on medical imaging. 2022;41(6):1533-46.

\bibitem{xue2022robust}
Xue C, Yu L, Chen P, Dou Q, Heng PA.
\newblock Robust medical image classification from noisy labeled data with global and local representation guided co-training.
\newblock IEEE Transactions on Medical Imaging. 2022;41(6):1371-82.

\bibitem{jiang2023label}
Jiang H, Gao M, Hu Y, Ren Q, Xie Z, Liu J.
\newblock Label-noise-tolerant medical image classification via self-attention and self-supervised learning.
\newblock arXiv preprint arXiv:230609718. 2023.

\bibitem{hu2021deep}
Hu K, Huang Y, Huang W, Tan H, Chen Z, Zhong Z, et~al.
\newblock Deep supervised learning using self-adaptive auxiliary loss for COVID-19 diagnosis from imbalanced CT images.
\newblock Neurocomputing. 2021;458:232-45.

\bibitem{lee2020fully}
Lee J, Prabhu D, Kolluru C, Gharaibeh Y, Zimin VN, Dallan LA, et~al.
\newblock Fully automated plaque characterization in intravascular OCT images using hybrid convolutional and lumen morphology features.
\newblock Scientific reports. 2020;10(1):2596.

\bibitem{kurita2023accurate}
Kurita Y, Meguro S, Tsuyama N, Kosugi I, Enomoto Y, Kawasaki H, et~al.
\newblock Accurate deep learning model using semi-supervised learning and Noisy Student for cervical cancer screening in low magnification images.
\newblock Plos one. 2023;18(5):e0285996.

\bibitem{fries2019weakly}
Fries JA, Varma P, Chen VS, Xiao K, Tejeda H, Saha P, et~al.
\newblock Weakly supervised classification of aortic valve malformations using unlabeled cardiac MRI sequences.
\newblock Nature communications. 2019;10(1):3111.

\bibitem{wong2022deep}
Wong DR, Tang Z, Mew NC, Das S, Athey J, McAleese KE, et~al.
\newblock Deep learning from multiple experts improves identification of amyloid neuropathologies.
\newblock Acta neuropathologica communications. 2022;10(1):66.

\bibitem{ashraf2022loss}
Ashraf M, Robles WRQ, Kim M, Ko YS, Yi MY.
\newblock A loss-based patch label denoising method for improving whole-slide image analysis using a convolutional neural network.
\newblock Scientific reports. 2022;12(1):1392.

\bibitem{ying2023covid}
Ying X, Liu H, Huang R.
\newblock COVID-19 chest X-ray image classification in the presence of noisy labels.
\newblock Displays. 2023;77:102370.

\bibitem{lopez2021learning}
L{\'o}pez-P{\'e}rez M, Amgad M, Morales-{\'A}lvarez P, Ruiz P, Cooper LA, Molina R, et~al.
\newblock Learning from crowds in digital pathology using scalable variational Gaussian processes.
\newblock Scientific reports. 2021;11(1):11612.

\bibitem{karimi2020learning}
Karimi D, Peters JM, Ouaalam A, Prabhu SP, Sahin M, Krueger DA, et~al.
\newblock Learning to detect brain lesions from noisy annotations.
\newblock In: 2020 IEEE 17th International Symposium on Biomedical Imaging (ISBI). IEEE; 2020. p. 1910-4.

\bibitem{dikici2022advancing}
Dikici E, Nguyen XV, Bigelow M, Ryu JL, Prevedello LM.
\newblock Advancing Brain Metastases Detection in T1-Weighted Contrast-Enhanced 3D MRI Using Noisy Student-Based Training.
\newblock Diagnostics. 2022;12(8):2023.

\bibitem{momeny2021learning}
Momeny M, Neshat AA, Hussain MA, Kia S, Marhamati M, Jahanbakhshi A, et~al.
\newblock Learning-to-augment strategy using noisy and denoised data: Improving generalizability of deep CNN for the detection of COVID-19 in X-ray images.
\newblock Computers in Biology and Medicine. 2021;136:104704.

\bibitem{jaiswal2022ros}
Jaiswal A, Ashutosh K, Rousseau JF, Peng Y, Wang Z, Ding Y.
\newblock Ros-kd: A robust stochastic knowledge distillation approach for noisy medical imaging.
\newblock In: 2022 IEEE International Conference on Data Mining (ICDM). IEEE; 2022. p. 981-6.

\bibitem{pulido2020semi}
Pulido JV, Guleria S, Ehsan L, Fasullo M, Lippman R, Mutha P, et~al.
\newblock Semi-supervised classification of noisy, gigapixel histology images.
\newblock In: 2020 IEEE 20th International Conference on Bioinformatics and Bioengineering (BIBE). IEEE; 2020. p. 563-8.

\bibitem{paul2021generalized}
Paul A, Shen TC, Lee S, Balachandar N, Peng Y, Lu Z, et~al.
\newblock Generalized zero-shot chest x-ray diagnosis through trait-guided multi-view semantic embedding with self-training.
\newblock IEEE Transactions on Medical Imaging. 2021;40(10):2642-55.

\bibitem{li2021pathal}
Li W, Li J, Wang Z, Polson J, Sisk AE, Sajed DP, et~al.
\newblock Pathal: An active learning framework for histopathology image analysis.
\newblock IEEE Transactions on Medical Imaging. 2021;41(5):1176-87.

\bibitem{gu2018reliable}
Gu Y, Shen M, Yang J, Yang GZ.
\newblock Reliable label-efficient learning for biomedical image recognition.
\newblock IEEE Transactions on Biomedical Engineering. 2018;66(9):2423-32.

\bibitem{zhang2021reur}
Zhang S, Yuan Z, Wang Y, Bai Y, Chen B, Wang H.
\newblock REUR: a unified deep framework for signet ring cell detection in low-resolution pathological images.
\newblock Computers in Biology and Medicine. 2021;136:104711.

\bibitem{gundel2021robust}
G{\"u}ndel S, Setio AA, Ghesu FC, Grbic S, Georgescu B, Maier A, et~al.
\newblock Robust classification from noisy labels: Integrating additional knowledge for chest radiography abnormality assessment.
\newblock Medical Image Analysis. 2021;72:102087.

\bibitem{algan2020deep}
Algan G, Ulusoy I, G{\"o}n{\"u}l {\c{S}}, Turgut B, Bakbak B.
\newblock Deep learning from small amount of medical data with noisy labels: A meta-learning approach.
\newblock arXiv preprint arXiv:201006939. 2020.

\bibitem{ghesu2019quantifying}
Ghesu FC, Georgescu B, Gibson E, Guendel S, Kalra MK, Singh R, et~al.
\newblock Quantifying and leveraging classification uncertainty for chest radiograph assessment.
\newblock In: Medical Image Computing and Computer Assisted Intervention--MICCAI 2019: 22nd International Conference, Shenzhen, China, October 13--17, 2019, Proceedings, Part VI 22. Springer; 2019. p. 676-84.

\bibitem{pham2021interpreting}
Pham HH, Le TT, Tran DQ, Ngo DT, Nguyen HQ.
\newblock Interpreting chest X-rays via CNNs that exploit hierarchical disease dependencies and uncertainty labels.
\newblock Neurocomputing. 2021;437:186-94.

\bibitem{irvin2019chexpert}
Irvin J, Rajpurkar P, Ko M, Yu Y, Ciurea-Ilcus S, Chute C, et~al.
\newblock Chexpert: A large chest radiograph dataset with uncertainty labels and expert comparison.
\newblock In: Proceedings of the AAAI conference on artificial intelligence. vol.~33; 2019. p. 590-7.

\bibitem{li2023learning}
Li J, Cao H, Wang J, Liu F, Dou Q, Chen G, et~al.
\newblock Learning Robust Classifier for Imbalanced Medical Image Dataset with Noisy Labels by Minimizing Invariant Risk.
\newblock In: International Conference on Medical Image Computing and Computer-Assisted Intervention. Springer; 2023. p. 306-16.

\bibitem{chen2022adaptive}
Chen H, Tan W, Li J, Guan P, Wu L, Yan B, et~al.
\newblock Adaptive Cross Entropy for ultrasmall object detection in Computed Tomography with noisy labels.
\newblock Computers in Biology and Medicine. 2022;147:105763.

\bibitem{xiang2023automatic}
Xiang J, Wang X, Wang X, Zhang J, Yang S, Yang W, et~al.
\newblock Automatic diagnosis and grading of Prostate Cancer with weakly supervised learning on whole slide images.
\newblock Computers in Biology and Medicine. 2023;152:106340.

\bibitem{del2023labeling}
Del~Amor R, Silva-Rodr{\'\i}guez J, Naranjo V.
\newblock Labeling confidence for uncertainty-aware histology image classification.
\newblock Computerized Medical Imaging and Graphics. 2023;107:102231.

\bibitem{calli2019handling}
Calli E, Sogancioglu E, Scholten ET, Murphy K, van Ginneken B.
\newblock Handling label noise through model confidence and uncertainty: application to chest radiograph classification.
\newblock In: Medical Imaging 2019: Computer-Aided Diagnosis. vol. 10950. SPIE; 2019. p. 289-96.

\bibitem{braun2022influence}
Braun J, Kornreich M, Park J, Pawar J, Browning J, Herzog R, et~al.
\newblock Influence Based Re-Weighing for Labeling Noise in Medical Imaging.
\newblock In: 2022 IEEE 19th International Symposium on Biomedical Imaging (ISBI). IEEE; 2022. p. 1-5.

\bibitem{jimenez2022curriculum}
Jim{\'e}nez-S{\'a}nchez A, Mateus D, Kirchhoff S, Kirchhoff C, Biberthaler P, Navab N, et~al.
\newblock Curriculum learning for improved femur fracture classification: Scheduling data with prior knowledge and uncertainty.
\newblock Medical Image Analysis. 2022;75:102273.

\bibitem{liu2021co}
Liu J, Li R, Sun C.
\newblock Co-correcting: noise-tolerant medical image classification via mutual label correction.
\newblock IEEE Transactions on Medical Imaging. 2021;40(12):3580-92.

\bibitem{hu2023fundus}
Hu T, Yang B, Guo J, Zhang W, Liu H, Wang N, et~al.
\newblock A fundus image classification framework for learning with noisy labels.
\newblock Computerized Medical Imaging and Graphics. 2023;108:102278.

\bibitem{shi2021correcting}
Shi P, Xin J, Zheng N.
\newblock Correcting Pseudo Labels with Label Distribution for Unsupervised Domain Adaptive Vulnerable Plaque Detection.
\newblock In: 2021 43rd Annual International Conference of the IEEE Engineering in Medicine \& Biology Society (EMBC). IEEE; 2021. p. 3225-8.

\bibitem{gao2023clinical}
Gao Z, Chen Y, Sun P, Liu H, Lu Y.
\newblock Clinical knowledge embedded method based on multi-task learning for thyroid nodule classification with ultrasound images.
\newblock Physics in Medicine \& Biology. 2023;68(4):045018.

\bibitem{gao2022bayesian}
Gao M, Feng X, Geng M, Jiang Z, Zhu L, Meng X, et~al.
\newblock Bayesian statistics-guided label refurbishment mechanism: Mitigating label noise in medical image classification.
\newblock Medical Physics. 2022;49(9):5899-913.

\bibitem{zhou2023refixmatch}
Zhou S, Tian S, Yu L, Wu W, Zhang D, Peng Z, et~al.
\newblock ReFixMatch-LS: reusing pseudo-labels for semi-supervised skin lesion classification.
\newblock Medical \& Biological Engineering \& Computing. 2023;61(5):1033-45.

\bibitem{zhu2023robust}
Zhu M, Zhang L, Wang L, Li D, Zhang J, Yi Z.
\newblock Robust co-teaching learning with consistency-based noisy label correction for medical image classification.
\newblock International Journal of Computer Assisted Radiology and Surgery. 2023;18(4):675-83.

\bibitem{javadi2022training}
Javadi G, Samadi S, Bayat S, Sojoudi S, Hurtado A, Eshumani W, et~al.
\newblock Training deep neural networks with noisy clinical labels: toward accurate detection of prostate cancer in US data.
\newblock International Journal of Computer Assisted Radiology and Surgery. 2022;17(9):1697-705.

\bibitem{chen2023combating}
Chen B, Ye Z, Liu Y, Zhang Z, Pan J, Zeng B, et~al.
\newblock Combating Medical Label Noise via Robust Semi-supervised Contrastive Learning.
\newblock In: International Conference on Medical Image Computing and Computer-Assisted Intervention. Springer; 2023. p. 562-72.

\bibitem{chen2023bomd}
Chen Y, Liu F, Wang H, Wang C, Liu Y, Tian Y, et~al.
\newblock Bomd: bag of multi-label descriptors for noisy chest x-ray classification.
\newblock In: Proceedings of the IEEE/CVF International Conference on Computer Vision; 2023. p. 21284-95.

\bibitem{boughorbel2018alternating}
Boughorbel S, Jarray F, Venugopal N, Elhadi H.
\newblock Alternating loss correction for preterm-birth prediction from ehr data with noisy labels.
\newblock arXiv preprint arXiv:181109782. 2018.

\bibitem{yang2023addressing}
Yang J, Triendl H, Soltan AA, Prakash M, Clifton DA.
\newblock Addressing Label Noise for Electronic Health Records: Insights from Computer Vision for Tabular Data.
\newblock medRxiv. 2023:2023-10.

\bibitem{murray2019automated}
Murray SG, Avati A, Schmajuk G, Yazdany J.
\newblock Automated and flexible identification of complex disease: building a model for systemic lupus erythematosus using noisy labeling.
\newblock Journal of the American Medical Informatics Association. 2019;26(1):61-5.

\bibitem{dhrangadhariya2023not}
Dhrangadhariya A, M{\"u}ller H.
\newblock Not so weak PICO: leveraging weak supervision for participants, interventions, and outcomes recognition for systematic review automation.
\newblock JAMIA open. 2023;6(1):ooac107.

\bibitem{li2021semi}
Li Z, Gan Z, Zhang B, Chen Y, Wan J, Liu K, et~al.
\newblock Semi-supervised noisy label learning for Chinese clinical named entity recognition.
\newblock Data Intelligence. 2021;3(3):389-401.

\bibitem{vazquez2022label}
V{\'a}zquez CG, Breuss A, Gnarra O, Portmann J, Madaffari A, Da~Poian G.
\newblock Label noise and self-learning label correction in cardiac abnormalities classification.
\newblock Physiological measurement. 2022;43(9):094001.

\bibitem{de2023stochastic}
de~Vos BD, Jansen GE, I{\v{s}}gum I.
\newblock Stochastic co-teaching for training neural networks with unknown levels of label noise.
\newblock Scientific reports. 2023;13(1):16875.

\bibitem{baghel2020automatic}
Baghel N, Dutta MK, Burget R.
\newblock Automatic diagnosis of multiple cardiac diseases from PCG signals using convolutional neural network.
\newblock Computer Methods and Programs in Biomedicine. 2020;197:105750.

\bibitem{vazquez2021two}
V{\'a}zquez CG, Breuss A, Gnarra O, Portmann J, Da~Poian G.
\newblock Two will do: CNN with asymmetric loss, self-learning label correction, and hand-crafted features for imbalanced multi-label ECG data classification.
\newblock In: 2021 Computing in Cardiology (CinC). vol.~48. IEEE; 2021. p. 1-4.

\bibitem{Ding2022LearningFA}
Ding C, Guo Z, Shah A, Clifford G, Rudin C, Hu X.
\newblock Learning From Alarms: A Robust Learning Approach for Accurate Photoplethysmography-Based Atrial Fibrillation Detection using Eight Million Samples Labeled with Imprecise Arrhythmia Alarms.
\newblock arXiv preprint arXiv:221103333. 2022.

\bibitem{hong2023semi}
Hong C, Liang L, Yuan Q, Cho K, Liao KP, Pencina MJ, et~al.
\newblock Semi-Supervised Calibration of Noisy Event Risk (SCANER) with Electronic Health Records.
\newblock Journal of Biomedical Informatics. 2023:104425.

\bibitem{ren2023ocrfinder}
Ren J, Liu Y, Zhu X, Wang X, Li Y, Liu Y, et~al.
\newblock OCRFinder: a noise-tolerance machine learning method for accurately estimating open chromatin regions.
\newblock Frontiers in Genetics. 2023;14:1184744.

\bibitem{tjandra2023leveraging}
Tjandra D, Wiens J.
\newblock Leveraging an Alignment Set in Tackling Instance-Dependent Label Noise.
\newblock In: Conference on Health, Inference, and Learning. PMLR; 2023. p. 477-97.

\bibitem{vernekar2022improving}
Vernekar S, Ayyar A, Rajagopalan A, Kumar B, Mishra VK.
\newblock Improving Medical Predictions with Label Noise Tolerant Classification.
\newblock In: 2022 4th International Conference on Advances in Computing, Communication Control and Networking (ICAC3N). IEEE; 2022. p. 765-9.

\bibitem{xu2020hybrid}
Xu J, Yang Y, Yang P.
\newblock Hybrid label noise correction algorithm for medical auxiliary diagnosis.
\newblock In: 2020 IEEE 18th International Conference on Industrial Informatics (INDIN). vol.~1. IEEE; 2020. p. 567-72.

\bibitem{brady2017error}
Brady AP.
\newblock Error and discrepancy in radiology: inevitable or avoidable?
\newblock Insights into imaging. 2017;8:171-82.

\bibitem{lu2019automated}
Lu H, Lee J, Ray S, Tanaka K, Bezerra HG, Rollins AM, et~al.
\newblock Automated stent coverage analysis in intravascular OCT (IVOCT) image volumes using a support vector machine and mesh growing.
\newblock Biomedical Optics Express. 2019;10(6):2809-28.

\bibitem{hekler2020effects}
Hekler A, Kather JN, Krieghoff-Henning E, Utikal JS, Meier F, Gellrich FF, et~al.
\newblock Effects of label noise on deep learning-based skin cancer classification.
\newblock Frontiers in Medicine. 2020;7:177.

\bibitem{campbell2017plus}
Campbell JP, Kalpathy-Cramer J, Erdogmus D, Ostmo S, Swan R, Sonmez K, et~al.
\newblock Plus disease in rop: why do experts disagree, and how can we improve diagnosis?
\newblock Journal of American Association for Pediatric Ophthalmology and Strabismus $\{$JAAPOS$\}$. 2017;21(4):e5-6.

\bibitem{cosentino2023inference}
Cosentino J, Behsaz B, Alipanahi B, McCaw ZR, Hill D, Schwantes-An TH, et~al.
\newblock Inference of chronic obstructive pulmonary disease with deep learning on raw spirograms identifies new genetic loci and improves risk models.
\newblock Nature Genetics. 2023:1-9.

\bibitem{ding2022impact}
Ding C, Pereira T, Xiao R, Lee RJ, Hu X.
\newblock Impact of label noise on the learning based models for a binary classification of physiological signal.
\newblock Sensors. 2022;22(19):7166.

\bibitem{pechenizkiy2006class}
Pechenizkiy M, Tsymbal A, Puuronen S, Pechenizkiy O.
\newblock Class noise and supervised learning in medical domains: The effect of feature extraction.
\newblock In: 19th IEEE symposium on computer-based medical systems (CBMS'06). IEEE; 2006. p. 708-13.

\bibitem{potapenko2022detection}
Potapenko I, Kristensen M, Thiesson B, Ilginis T, Lykke~S{\o}rensen T, Nouri~Hajari J, et~al.
\newblock Detection of oedema on optical coherence tomography images using deep learning model trained on noisy clinical data.
\newblock Acta Ophthalmologica. 2022;100(1):103-10.

\bibitem{khanal2023investigating}
Khanal B, Hasan SK, Khanal B, Linte CA.
\newblock Investigating the impact of class-dependent label noise in medical image classification.
\newblock In: Medical Imaging 2023: Image Processing. vol. 12464. SPIE; 2023. p. 728-33.

\bibitem{samala2020generalization}
Samala RK, Chan HP, Hadjiiski LM, Helvie MA, Richter CD.
\newblock Generalization error analysis for deep convolutional neural network with transfer learning in breast cancer diagnosis.
\newblock Physics in Medicine \& Biology. 2020;65(10):105002.

\bibitem{buttner2023impact}
B{\"u}ttner M, Schneider L, Krasowski A, Krois J, Feldberg B, Schwendicke F.
\newblock Impact of Noisy Labels on Dental Deep Learning—Calculus Detection on Bitewing Radiographs.
\newblock Journal of Clinical Medicine. 2023;12(9):3058.

\bibitem{jang2020assessment}
Jang R, Kim N, Jang M, Lee KH, Lee SM, Lee KH, et~al.
\newblock Assessment of the robustness of convolutional neural networks in labeling noise by using chest X-ray images from multiple centers.
\newblock JMIR medical informatics. 2020;8(8):e18089.

\bibitem{petersen2023path}
Petersen E, Holm S, Ganz M, Feragen A.
\newblock The path toward equal performance in medical machine learning.
\newblock Patterns. 2023;4(7).

\bibitem{liu2015classification}
Liu T, Tao D.
\newblock Classification with noisy labels by importance reweighting.
\newblock IEEE Transactions on pattern analysis and machine intelligence. 2015;38(3):447-61.

\bibitem{patrini2017making}
Patrini G, Rozza A, Krishna~Menon A, Nock R, Qu L.
\newblock Making deep neural networks robust to label noise: A loss correction approach.
\newblock In: Proceedings of the IEEE conference on computer vision and pattern recognition; 2017. p. 1944-52.

\bibitem{goldberger2016training}
Goldberger J, Ben-Reuven E.
\newblock Training deep neural-networks using a noise adaptation layer.
\newblock In: International conference on learning representations; 2016. .

\end{thebibliography}












\end{document}